\documentclass[final]{clv2025}

\usepackage{amsmath}
\usepackage{amssymb}
\usepackage{booktabs}
\usepackage{graphicx}
\usepackage{microtype}
\hypersetup{
  hidelinks,
  pdftitle={Multi-Label Agreement: A Flexible Framework for Inter-Annotator Reliability},
  pdfauthor={Tim Menzner and Jochen L. Leidner},
  pdfsubject={Preprint on inter-annotator agreement for multi-label annotation},
  pdfkeywords={inter-annotator agreement, multi-label annotation, Fleiss kappa, Krippendorff alpha, Gwet AC1, prevalence}
}

\jname{Preprint}
\jinfo{October 2026}
\jvol{}
\jnum{}
\jyear{2026}
\dochead{Preprint}
\renewcommand{\footmark}{Preprint. This version has not undergone peer review.}

\runningtitle{Multi-Label Agreement}
\runningauthor{Menzner and Leidner}

\begin{document}

\title{Multi-Label Agreement: A Flexible Framework for Inter-Annotator Reliability}

\author{Tim Menzner$^{1}$, Jochen L. Leidner$^{2}$}

\affilblock{
    \affil{Center for Responsible Artificial Intelligence (CRAI), Coburg University of Applied Sciences\\\quad \email{tim.menzner@hs-coburg.de}}
    \affil{Center for Responsible Artificial Intelligence (CRAI), Coburg University of Applied Sciences\\\quad \email{leidner@acm.org}}
}

\maketitle

\begin{abstract}
  Reliable evaluation of inter-annotator agreement (IAA) is important for assessing the consistency of annotation processes. While established IAA metrics such as Cohen's kappa, Fleiss' kappa, and Krippendorff's alpha are commonly applied in single-label settings, multi-label tasks require additional decisions about how full label sets, joint selections, and joint omissions should be compared.

  In this work, we present Multi-Label Agreement (MLA), a flexible framework for evaluating agreement in multi-label annotation tasks. MLA preserves the full binary label vectors assigned to each item and uses average pairwise Hamming agreement, thereby accounting for both joint selections and joint omissions. We show that the pooled-marginal variant can be expressed as a chance-disagreement-weighted average of label-wise Fleiss/Scott-type coefficients. Apart from the finite-sample convention used for expected disagreement, it corresponds to Krippendorff's alpha with normalized Hamming distance. The framework supports pooled-marginal, Gwet AC1, and fixed-baseline normalizations, yields item-level, label-level, and corpus-level results, and includes a conditional minimum-agreement diagnostic. Through a worked example and Monte Carlo simulations, we show how the resulting values depend on label prevalence, the declared label inventory, and the selected model of expected agreement. We also provide a Python implementation. MLA thus offers a transparent and interpretable framework for multi-label agreement analysis without claiming that one normalization is universally preferable.
\end{abstract}

\section{Introduction}

To assess the reproducibility of an annotation process, annotation studies commonly report \emph{inter-annotator agreement}, also known as inter-coder or inter-rater agreement \cite{Carletta:1996:CL,artstein-poesio-2008-survey,Leidner:2022:NLDB}. Agreement does not by itself prove that the annotations are valid or that they form an upper bound for a downstream classifier, but it can reveal unclear guidelines, inconsistent use of categories, and systematic differences between annotators.

This is done by having a set of human subjects annotate a shared set of data items so their individual decisions can be compared. After all annotators annotate the same piece of data and inter-annotator agreement (IAA) is established, annotators may each work on their own data subset without overlap to save time and money.

Factors that hurt IAA include unspecified, ambiguous, or differently interpreted annotation guidelines as well as human performance effects like lapses of concentration. A good annotation guideline codebook describes each class clearly and provides examples and counterexamples to differentiate it from other classes. Typically, writing the codebook requires several iterations, with multiple updates following each round of pre-trial annotations \cite{Artstein:2017,Leidner:2022:NLDB}. At the same time, disagreement is not always annotation noise. In subjective tasks it may reflect genuine differences in interpretation and can itself be useful information \cite{plank-etal-2014-linguistically,uma-etal-2021-disagreement}.

Tasks may be inherently ambiguous given a particular set of categories, or different annotators may display divergent annotation behavior even when using the same codebook, due, for example, to concentration issues.

Depending on the annotation setup, several established metrics can be used to assess IAA. With two annotators and one categorical label for each item, Cohen's $\kappa$ \citep{cohen1960} uses annotator-specific marginal distributions. Scott's $\pi$ and Fleiss' $\kappa$ instead use pooled marginals, with Fleiss' formulation accommodating more than two annotations per item \citep{fleiss1971,di-eugenio-glass-2004-squibs,artstein-poesio-2008-survey}. Krippendorff's $\alpha$ \citep{krippendorff2011,Krippendorff:2013} supports any number of annotators, missing data, and user-defined distance functions. This last feature also permits set-valued or vector-valued annotations.

For a formal description of a multi-label annotation task, let \(N\) be the number of data items \(i\) to be annotated by \(n\) annotators, each of whom assigns a total of \(k\) category labels (assigning in this case also includes not assigning, as the absence of a label is also information). For each item and annotator, we therefore do not have a single decision but a set of decisions \(L\) with \(L \in \{0, 1\}^k\).

Most common IAA metrics, however, such as the ones mentioned above, assume that when there are \(k\) possible categories for each item \(i\), each annotator must choose one of the labels. Thus, for each item and annotator, you end up with \(\ell_{i,c} \in \{1, 2, \dots, k\}\), where \(\ell_{i,c}\) denotes the single category label selected by annotator (coder) \(c\) for item \(i\).

There are at least two natural ways to proceed. One can treat every label as a separate binary annotation problem and combine the resulting statistics, or treat the entire label set as one structured value and select a distance between two sets. These approaches are not necessarily different: normalized Hamming distance is additive over labels, and therefore reduces the vector comparison to a particular weighting of label-wise comparisons.

This paper makes that relationship explicit. We call the resulting framework \emph{Multi-Label Agreement} (MLA), retaining the terminology used in the accompanying software. MLA is not a mathematically independent coefficient family; instead, our contributions are (i) an exact decomposition into label-wise coefficients, (ii) the connection to Krippendorff's $\alpha$ with Hamming distance, (iii) a conditional lower bound showing how far agreement can fall for the observed prevalences, and (iv) a practical comparison of alternative chance terms and set similarities.

\section{Related Work} \label{related_work}

\subsection{Past Work on Multi-Label IAA}

Multi-label agreement has been approached both through dedicated coefficients and through general weighted agreement frameworks. The latter perspective is particularly important because Krippendorff's $\alpha$ can treat a label set as a structured value and use a distance such as Hamming, Jaccard, or MASI \citep{artstein-poesio-2008-survey,passonneau-2006-masi}. Recent annotation systems consequently use $\alpha$ with MASI distance for multi-label monitoring \citep{alanno-2023}.

\citet{rosenberg-binkowski-2004-augmenting} proposed an augmentation of Cohen's kappa, treating multiple labels not as a legitimate outcome of a task but rather a problem arising from annotator uncertainty, following the assumption that there has to be one true label and annotators just struggle to agree which one it is. The way it is calculated essentially boils down to the kappa for a scenario where coders were only allowed to give one label and had randomly selected one of multiple provided labels. Thus, this metric has been designed to deflate with the number of labels used, a higher number of labels automatically leads to a lower score. This is not really suitable for an annotation task where an item could have several true labels. 

\citet{marchal-etal-2022-establishing} propose a computational approach that uses bootstrapping to estimate chance agreement in multi-label data, while assessing agreement only over the intersection of annotators' label sets and deliberately ignoring labels that were never assigned to an item. While this is a defensible design choice, the simulated (rather than analytic) chance term can be more difficult to interpret, and discarding the information conveyed by joint non-selections means that agreement on a label's absence is treated as if the label had never been available for that item in the first place.

MASI was introduced by \citet{passonneau-2006-masi} and combines the Jaccard index with a penalty factor based on the degree of overlap, ranging from no penalty for an exact match, to a small penalty for a subset match, to a higher penalty for non-subset overlap between label sets. MASI itself is a distance or similarity, not a chance-corrected coefficient; it can be reported directly or supplied as the distance function in a coefficient such as $\alpha$.

\citet{bhowmick-etal-2008-agreement} introduce the Am coefficient, a $\kappa$-style statistic that extends pairwise agreement to the multi-label setting by treating every unordered pair of categories as a separate nominal variable. For each item, annotators' three possible decisions for a category pair, (0,0), (1,1), or (0,1)/(1,0), are compared; the observed overlap is then normalized with a chance term derived from the product of the individual annotators' marginal probabilities. Because it averages over all pairs, it implicitly weights errors between frequently co-occurring labels more heavily, yields only a single corpus-wide score, and inherits $\kappa$'s sensitivity to skewed prevalence while further scaling quadratically in the number of labels.

Related work on learning from disagreement and soft evaluation asks a different but connected question: whether the distribution of annotator judgements should be preserved rather than reduced to a single gold label \citep{uma-etal-2021-disagreement,rizzi-etal-2024-soft,harju-mesaros-2023}. Multi-label confusion matrices and tensors likewise provide detailed classification diagnostics but do not themselves model chance agreement \citep{heydarian-etal-2022-mlcm,krstinic-etal-2024-tensor}. These lines of work reinforce a central point of the present paper: a single corpus-level coefficient should be accompanied by diagnostics that preserve where and how disagreement occurs.

\subsection{The Ongoing Chance Correction Debate} \label{discussion}

Many agreement coefficients have the form $(A_o-A_e)/(1-A_e)$, where $A_o$ is observed agreement and $A_e$ is a model-dependent baseline. The word ``chance'' can obscure that $A_e$ is not observed independently of the data: it follows from a particular model of how annotations would be allocated in the absence of the agreement of interest \citep{di-eugenio-glass-2004-squibs,artstein-poesio-2008-survey}.

For a binary label $j$, let $p_j$ denote the pooled proportion of positive decisions across annotators and items. The corresponding baseline $p_j^2+(1-p_j)^2$ is Fleiss/Scott-type because it uses a pooled marginal. It should not be called Cohen's $\kappa$, whose two-annotator model keeps the annotators' marginal distributions separate.

Kappa-like coefficients can become small in situations with high raw agreement and skewed prevalence. Gwet's AC1 \citep{gwet2001handbook} uses a different baseline and often remains higher in such situations. PABAK \citep{Byrt1993}, also known in this binary form as the Brennan--Prediger coefficient, fixes the baseline at $0.5$.

Konstantinidis et al.\ \citep{konstantinidis2022empirical} provide simulation evidence that Gwet's AC1 often attains lower sampling variance than several kappa estimators in multiple-rater, binary-outcome settings and therefore recommend AC1 when skewed prevalences and small sample sizes make kappa unstable.  

Zec et al.\ \citep{zec2017high_agreement_high_prevalence} highlight the classic "kappa paradox", cases of high raw agreement but extreme category prevalence where Cohen's \(\kappa\) can appear deceptively low, and show empirical examples in which AC1 (and related measures) avoid that particular interpretive pitfall.  
Conversely, Vach and Gerke \citep{vach2023gwets_ac1_not_substitute} stress that AC1 is not a drop-in replacement: substituting AC1 for $\kappa$ changes the underlying model rather than repairing the same estimand. We therefore treat these normalizations as sensitivity analyses. Their values answer different questions and should not be selected after looking at which one gives the most favourable result.

\section{Multi-Label Agreement as a Decomposable Construction} \label{MultiLabel}

\subsection{Observed agreement}

Let $L_{c,i}=(\ell_{c,i,1},\ldots,\ell_{c,i,k})\in\{0,1\}^k$ be the vector assigned to item $i$ by annotator $c$. The normalized Hamming distance and its corresponding agreement are
\[
d_H(L_{c,i},L_{c',i})
=\frac{1}{k}\sum_{j=1}^{k}\mathbf{1}(\ell_{c,i,j}\neq\ell_{c',i,j}),
\qquad
s_H=1-d_H.
\]
Thus, a joint selection $(1,1)$ and a joint omission $(0,0)$ both count as agreement. With $h=\binom{n}{2}$ annotator pairs, the item-level and corpus-level observed agreements are
\begin{align}
A_{o,i}
&=\frac{1}{h}\sum_{1\leq c<c'\leq n}s_H(L_{c,i},L_{c',i}),\\
A_o
&=\frac{1}{N}\sum_{i=1}^{N}A_{o,i}.
\end{align}
For label $j$ alone, define
\[
A_{o,j}=\frac{1}{Nh}\sum_{i=1}^{N}\sum_{c<c'}
\mathbf{1}(\ell_{c,i,j}=\ell_{c',i,j}).
\]
Because Hamming agreement is additive, $A_o=k^{-1}\sum_j A_{o,j}$. This equality is useful but also marks a limitation: the construction compares the components of the vectors independently and does not retain information about dependencies or co-occurrence patterns between labels.

\subsection{Alternative baseline models}

For binary label $j$, its pooled prevalence is
\[
p_j=\frac{\sum_{i=1}^{N}\sum_{c=1}^{n}\ell_{c,i,j}}{nN}.
\]
The pooled-marginal (Fleiss/Scott-type) agreement baseline is
\[
A_{e,j}^{(F)}=p_j^2+(1-p_j)^2,
\qquad
A_e^{(F)}=\frac{1}{k}\sum_{j=1}^{k}A_{e,j}^{(F)}.
\]
For comparison, the binary AC1 chance term and the fixed PABAK chance term are
\begin{align}
A_{e,j}^{(AC1)}&=2p_j(1-p_j),
&A_e^{(AC1)}&=\frac{1}{k}\sum_{j=1}^{k}A_{e,j}^{(AC1)},\\
A_{e,j}^{(PB)}&=0.5,
&A_e^{(PB)}&=0.5.
\end{align}
For binary decisions $A_{e,j}^{(AC1)}=1-A_{e,j}^{(F)}$, but the two quantities are used as baselines in different coefficient models and must not be interpreted as interchangeable estimates of the same null process.

For $m\in\{F,AC1,PB\}$, the aggregate coefficient is
\begin{equation}
A_{ml}^{(m)}=\frac{A_o-A_e^{(m)}}{1-A_e^{(m)}}.
\label{eq:aggregate}
\end{equation}
It equals one for perfect observed agreement. A value of zero means that $A_o$ equals the selected baseline; what that statement means depends on $m$. Item-level values can be obtained by replacing $A_o$ with $A_{o,i}$ while keeping the corpus-level baseline, but these are diagnostics relative to the corpus and not independent reliability estimates.

The raw prevalence
\[
p_r=\frac{1}{k}\sum_{j=1}^{k}p_j
\]
is useful as a compact summary, but it does not replace the label-wise prevalences because very different vectors $(p_1,\ldots,p_k)$ can have the same $p_r$.

\subsection{Exact relation to label-wise coefficients}

Let
\[
A_{j}^{(m)}=\frac{A_{o,j}-A_{e,j}^{(m)}}{1-A_{e,j}^{(m)}}
\]
be the coefficient for label $j$, whenever its denominator is non-zero, and let
$\mathcal J_m=\{j:1-A_{e,j}^{(m)}>0\}$. Labels outside $\mathcal J_m$
contribute zero to both the numerator and denominator below. Substituting
$A_o=k^{-1}\sum_jA_{o,j}$ and $A_e^{(m)}=k^{-1}\sum_jA_{e,j}^{(m)}$
into Equation~\ref{eq:aggregate} gives, provided that the aggregate denominator
is non-zero,
\begin{equation}
A_{ml}^{(m)}
=\frac{\sum_{j\in\mathcal J_m}\left(1-A_{e,j}^{(m)}\right)A_j^{(m)}}
{\sum_{j\in\mathcal J_m}\left(1-A_{e,j}^{(m)}\right)}.
\label{eq:weighted}
\end{equation}
The aggregate is therefore a weighted average of label-wise coefficients, not a new functional family. Under the pooled-marginal baseline, the weights are $2p_j(1-p_j)$, so labels with extreme prevalence receive less weight. Under AC1 they are $p_j^2+(1-p_j)^2$, which gives relatively more weight to extreme labels. Under PABAK all weights are equal. These implicit weights should be reported because they determine what the corpus-level number summarizes.

\subsection{Relation to Krippendorff's alpha}

The same construction can be written as a distance-based coefficient. Let $D_o=1-A_o$ be observed normalized Hamming disagreement. Drawing two label vectors independently from the pooled empirical distribution gives
\[
D_e^{(\infty)}=\frac{1}{k}\sum_{j=1}^{k}2p_j(1-p_j)
=1-A_e^{(F)}.
\]
Consequently,
\[
A_{ml}^{(F)}=1-\frac{D_o}{D_e^{(\infty)}}.
\]
This has the form of Krippendorff's $\alpha$ with normalized Hamming distance. Standard $\alpha$ draws two distinct values without replacement from the $M=nN$ observed vectors and, for complete balanced data, uses
\[
D_e^{(\alpha)}=\frac{M}{M-1}D_e^{(\infty)},
\qquad
\alpha_H=1-\frac{D_o}{D_e^{(\alpha)}}.
\]
Thus the pooled-marginal MLA coefficient and Hamming $\alpha$ differ only by this finite-sample convention and converge as $M$ grows. Krippendorff's framework is more general because it also provides explicit handling of missing data and permits other distances. For example, Jaccard or MASI distance deemphasizes joint omissions and instead focuses on overlap between selected labels.

\subsection{Minimum feasible agreement conditional on prevalence}

With more than two annotators, zero pairwise agreement is generally impossible for a binary decision. Without fixing the prevalence, the smallest possible pairwise observed agreement for a label is
\[
A_{o,\min}=\begin{cases}
\dfrac{n-2}{2(n-1)}, & n\text{ even},\\[4pt]
\dfrac{n-1}{2n}, & n\text{ odd}.
\end{cases}
\]
A more informative diagnostic keeps the observed number of positive decisions fixed. Let
\[
T_j=\sum_{i=1}^{N}\sum_{c=1}^{n}\ell_{c,i,j},
\qquad
x_{i,j}=\sum_{c=1}^{n}\ell_{c,i,j}.
\]
For an item with $x$ positive and $n-x$ negative decisions, the number of agreeing annotator pairs is
\[
f(x)=\binom{x}{2}+\binom{n-x}{2}.
\]
The function $f$ is convex on the integers. If two item counts differ by at least two, moving one positive decision from the larger to the smaller count cannot increase their combined pair agreement. Repeating this exchange shows that, subject to $\sum_i x_{i,j}=T_j$, agreement is minimized when all item counts differ by at most one.

Set
\[
m_j=\left\lfloor\frac{T_j}{N}\right\rfloor,
\qquad
t_j=T_j-m_jN.
\]
Then $N-t_j$ items receive $m_j$ positives and $t_j$ items receive $m_j+1$ positives. The conditional lower bound is therefore
\begin{equation}
A_{o,\min,j}^{(p_j)}
=\frac{(N-t_j)f(m_j)+t_jf(m_j+1)}{N\binom{n}{2}},
\label{eq:floor}
\end{equation}
where the second term is omitted when $t_j=0$. Averaging over labels and applying the same normalization yields
\[
A_{o,\min}^{(\mathbf p)}=\frac{1}{k}\sum_{j=1}^{k}A_{o,\min,j}^{(p_j)},
\qquad
A_{ml,\min}^{(m)}
=\frac{A_{o,\min}^{(\mathbf p)}-A_e^{(m)}}{1-A_e^{(m)}}.
\]
This is a feasible-range diagnostic, not another chance model and not part of the observed agreement. It explains why a coefficient may be unable to attain strongly negative values once the number of annotators and empirical marginals are fixed.

\subsection{Interpreting Prevalence} \label{prevalence}

When the raw prevalence $p_r$ is near 0, almost every coder assigns 0 to almost every label. Because a $(0,0)$ comparison counts as agreement, $A_o$ can become high even when coders rarely coincide on the scarce positive cases. This matters in tasks where recognizing an occasional positive label is the actual challenge. Take the rabbit--duck illusion: coders may dispute whether a rabbit, a duck, or both are present, yet they will unanimously agree that other animals are absent. Expanding the inventory with irrelevant animals therefore inflates Hamming agreement without improving reliability on the labels of interest.

A low $p_r$ is not automatically a problem. In some applications positives are genuinely rare and certifying their absence is an important part of the task. Think of aircraft-maintenance engineers who tick possible defects on an inspection checklist: almost all boxes will, and should, remain unticked. Here, joint omissions may be meaningful. The relevant question is therefore not whether negative matches should always count, but whether all labels were genuinely considered and whether confirming their absence has comparable value to confirming their presence.

The mirror image occurs when $p_r$ is close to 1: $A_o$ is then dominated by $(1,1)$ matches. Whether this is informative again depends on the task. A single global prevalence is insufficient for making this judgement, so the label-wise $p_j$, $A_{o,j}$, positive agreement, and negative agreement should be inspected alongside a corpus-level coefficient.

Natural ambiguity also has to be separated from annotation error. Genre annotation, for example, may legitimately produce different but defensible label sets. Such disagreement should often be preserved for later analysis rather than hidden behind a reliability threshold \citep{plank-etal-2014-linguistically,uma-etal-2021-disagreement}.

No normalization resolves this domain decision automatically. The pooled-marginal coefficient asks how much observed Hamming agreement exceeds that implied by the empirical label marginals. AC1 uses a different baseline that remains less punitive at extreme prevalence, but can consequently be positive even when annotators draw independently from the same skewed Bernoulli distribution. PABAK fixes a balanced baseline and has the same issue under skewed independent decisions. These are properties of the respective models, not evidence that one coefficient has recovered the uniquely correct amount of reliability. We therefore recommend pre-specifying the primary model and reporting $A_o$, label-wise prevalences, and at least one sensitivity analysis.

\subsection{Example}

To demonstrate the calculations, consider the following subsample from an experiment where three annotators were asked to annotate bias in sentences from news articles using a pre-specified taxonomy of 31 types. Table~\ref{tab:1} shows the selected types and the corresponding vectors. This sample is far too small for a substantive reliability conclusion; it is only a worked example.

\begin{table}[htbp]
\centering
\small
\begin{tabular}{|c|p{3cm}|p{3cm}|p{3cm}|}
\hline
\textbf{Sentence} & \textbf{Annotator 1} & \textbf{Annotator 2} & \textbf{Annotator 3} \\
\hline
1 & 
Faulty Analogy Bias \newline
[0,0,0,0,0,0,0,0,0,0,\newline 
0,0,0,1,0,0,0,0,0,0,\newline 
0,0,0,0,0,0,0,0,0,0,0] & 
Faulty Analogy Bias \newline
[0,0,0,0,0,0,0,0,0,0,\newline 
0,0,0,1,0,0,0,0,0,0,\newline 
0,0,0,0,0,0,0,0,0,0,0] & 
Faulty Analogy Bias \newline
[0,0,0,0,0,0,0,0,0,0,\newline 
0,0,0,1,0,0,0,0,0,0,\newline 
0,0,0,0,0,0,0,0,0,0,0] \\
\hline
2 & 
Political Bias, Projection Bias \newline
[0,0,0,0,0,0,0,0,0,0,\newline 
0,0,0,0,0,0,0,0,0,0,\newline 
0,1,1,0,0,0,0,0,0,0,0] & 
Speculation Bias, Unsubstantiated Claims Bias, Word Choice Bias \newline
[0,0,0,0,0,0,0,0,0,0,\newline 
0,0,0,0,0,0,0,0,0,0,\newline 
0,0,0,0,0,0,1,0,1,0,1] & 
Unsubstantiated Claims Bias, Word Choice Bias \newline
[0,0,0,0,0,0,0,0,0,0,\newline 
0,0,0,0,0,0,0,0,0,0,\newline 
0,0,0,0,0,0,0,0,1,0,1] \\
\hline
3 & 
Ambiguous Attribution Bias, Cherry Picking Bias, Political Bias \newline
[0,1,0,0,1,0,0,0,0,0,\newline 
0,0,0,0,0,0,0,0,0,0,\newline 
0,1,0,0,0,0,0,0,0,0,0] & 
Anecdotal Evidence Bias \newline
[0,0,1,0,0,0,0,0,0,0,\newline 
0,0,0,0,0,0,0,0,0,0,\newline 
0,0,0,0,0,0,0,0,0,0,0] & 
Ambiguous Attribution Bias \newline
[0,1,0,0,0,0,0,0,0,0,\newline 
0,0,0,0,0,0,0,0,0,0,\newline 
0,0,0,0,0,0,0,0,0,0,0] \\
\hline
4 & 
Intergroup Bias, Opinionated Bias, Political Bias, Word Choice Bias \newline
[0,0,0,0,0,0,0,0,0,0,\newline 
0,0,0,0,0,0,1,0,0,0,\newline 
1,1,0,0,0,0,0,0,0,0,1] & 
Ambiguous Attribution Bias, Intergroup Bias, Political Bias, Projection Bias, Unsubstantiated Claims Bias \newline
[0,1,0,0,0,0,0,0,0,0,\newline 
0,0,0,0,0,0,1,0,0,0,\newline 
0,1,1,0,0,0,0,0,1,0,0] & 
Ambiguous Attribution Bias, Opinionated Bias, Projection Bias, Word Choice Bias \newline
[0,1,0,0,0,0,0,0,0,0,\newline 
0,0,0,0,0,0,0,0,0,0,\newline 
1,0,1,0,0,0,0,0,0,0,1] \\
\hline
\end{tabular}
\caption{Assigned bias labels and corresponding 31-bit vectors by sentence and annotator.}
\label{tab:1}
\end{table}

We take \(N=4\) sentences, \(k=31\) categories, \(n=3\) annotators (\(h=\binom{3}{2}=3\)).  

\vspace{1ex}
For each sentence \(i\), compute
\[
S_{\mathcal{L}_i}
=\sum_{c<c'} I(L_{c,i},L_{c',i}),\quad
A_{o_i}=\frac{S_{\mathcal{L}_i}}{31\cdot 3}
\]
\begin{align*}
i=1:\quad & S=31+31+31=93
\quad\Rightarrow\;
A_{o_1}=93/93=1.000,\;\\
i=2:\quad & S=26+27+30=83
\quad\Rightarrow\;
A_{o_2}=83/93\approx0.8925,\;\\
i=3:\quad & S=27+29+29=85
\quad\Rightarrow\;
A_{o_3}=85/93\approx0.9140,\;\\
i=4:\quad & S=26+27+26=79
\quad\Rightarrow\;
A_{o_4}=79/93\approx0.8495,\;
\end{align*}

For the pooled-marginal and AC1 baseline terms we obtain

\[
A_{e}^{(F)}=\frac{27.5278}{31}\approx0.8880,\qquad
A_{e}^{(ac1)}=1-A_{e}^{(F)}\approx0.1120 .
\]

Hence the overall multi-label agreements under the three methods are  

\begin{align*}
A_{ml}^{(F)}
&= \frac{\tfrac{1+0.8925+0.9140+0.8495}{4}-0.8880}{1-0.8880}
\approx 0.2320,\\
A_{ml}^{(ac1)}&\approx 0.9031,
&A_{ml}^{(pb)}&\approx 0.8280.
\end{align*}

 As for the raw prevalence, across all \(12=4\times3\) decisions per label there are \(28\) positive assignments, so  

\[
p_r = \frac{1}{k}\sum_{j=1}^{k} p_j
     =\frac{28}{12\cdot31}
     =\frac{7}{93}
     \approx 0.0753 .
\]

The minimum observed agreements given the empirical prevalence based on the total number of 1s are \\

\begin{tabular}{r|c|c}
\text{total "1"s for a label} & \#\,\text{labels} & \(A_{o_{\min},j}^{p_j}\) \\\hline
0 & 20 & \(1.0000\) \\
1 &  3 & \(5/6 = 0.8333\) \\
2 &  2 & \(2/3 = 0.6667\) \\
3 &  3 & \(1/2 = 0.5000\) \\
4 &  3 & \(1/3 = 0.3333\) \\\hline
\end{tabular} \\

Averaging over all 31 labels gives  

\[
A_{o_{\min}}^{p_j}=
\frac{20\!\cdot\!1 + 3\!\cdot\!\frac56 + 2\!\cdot\!\frac23 + 3\!\cdot\!\frac12 + 3\!\cdot\!\frac13}{31}
=\frac{79}{93}\approx0.8495 .
\]

This sets the three normalized floors to

\begin{equation}
\begin{aligned}
A_{ml,\min}^{(F)}
&= \frac{\tfrac{79}{93}-0.8880}{1-0.8880} \approx -0.3440,\\
A_{ml,\min}^{(ac1)}
&= \frac{\tfrac{79}{93}-0.1120}{1-0.1120} \approx 0.8305,\\
A_{ml,\min}^{(pb)}
&= \frac{\tfrac{79}{93}-0.5000}{0.5000} \approx 0.6989.
\end{aligned}
\end{equation}

This is a scenario with very low prevalence. The pooled-marginal coefficient is low because most raw agreement is also implied by the empirical marginals. AC1 and PABAK are high, but their conditional floors are high as well: given the observed number of positive decisions, these coefficients cannot approach $-1$ and in the AC1 case cannot even approach zero. This is why a value without its baseline and feasible range is easy to overinterpret.

Twenty of the 31 pre-specified categories are never selected in the example. Table~\ref{tab:label-universe} shows the effect of removing them post hoc. The pooled-marginal coefficient and its floor remain unchanged because constant labels have weight $1-A_{e,j}^{(F)}=0$ in Equation~\ref{eq:weighted}. AC1 and PABAK change substantially because constant labels receive non-zero weight under these models. Neither label universe is automatically correct: the full inventory is appropriate if all annotators genuinely considered all 31 labels, whereas post-hoc removal changes the estimand and can hide the effect of an unnecessarily broad label inventory.

\begin{table}[htbp]
\centering
\begin{tabular}{lccccc}
\toprule
Label universe & $A_o$ & $A_{ml}^{(F)}$ & $A_{ml}^{(ac1)}$ & $A_{ml}^{(pb)}$ & $A_{o,\min}^{(\mathbf p)}$\\
\midrule
All 31 labels & 0.914 & 0.232 & 0.903 & 0.828 & 0.849\\
11 selected labels & 0.758 & 0.232 & 0.646 & 0.515 & 0.576\\
\bottomrule
\end{tabular}
\caption{Effect of the declared label universe in the worked example. Removing never-selected labels leaves the pooled-marginal coefficient unchanged but changes the AC1 and fixed-baseline summaries.}
\label{tab:label-universe}
\end{table}

Table~\ref{tab:perlabel} reports, for every category that received at least one positive assignment, its prevalence, observed agreement, and the three label-wise coefficients. Their variation highlights why a corpus-level number should not be reported alone.

To put these results into perspective, the same example was also used to calculate the measures mentioned in Section~\ref{related_work} ($B$ gives the number of bootstrap runs):
\begin{align*}
\text{Augmented }\kappa &=0.253, &
\text{boot-F1 }(B=1000)&=0.366,\\
\text{MASI}&=0.509, &
A_m&=0.281.
\end{align*}

These values are not estimates of one common quantity. Augmented $\kappa$ and $A_m$ apply their own chance models, boot-F1 ignores joint negative decisions, and the directly reported MASI similarity is not chance-corrected. Their disagreement is therefore informative: different summaries reward different aspects of the same annotations.

\begin{table}[h!]
\centering
\begin{tabular}{r|c|c|c|c|c}
\textbf{label $j$} & $p_j$ & $A_{o,j}$ &
$A_{ml,j}^{(F)}$ & $A_{ml,j}^{(ac1)}$ & $A_{ml,j}^{(pb)}$\\\hline
 2 (Ambiguous Attribution Bias) & 0.333 & 0.667 & 0.250 & 0.400 & 0.333\\
 3 (Anecdotal Evidence Bias) & 0.083 & 0.833 & $-0.091$ & 0.803 & 0.667\\
 5 (Cherry Picking Bias) & 0.083 & 0.833 & $-0.091$ & 0.803 & 0.667\\
14 (Faulty Analogy Bias) & 0.250 & 1.000 & 1.000 & 1.000 & 1.000\\
17 (Intergroup Bias) & 0.167 & 0.833 & 0.400 & 0.769 & 0.667\\
21 (Opinionated Bias) & 0.167 & 0.833 & 0.400 & 0.769 & 0.667\\
22 (Political Bias) & 0.333 & 0.500 & $-0.125$ & 0.100 & 0.000\\
23 (Projection Bias) & 0.250 & 0.667 & 0.111 & 0.467 & 0.333\\
27 (Speculation Bias) & 0.083 & 0.833 & $-0.091$ & 0.803 & 0.667\\
29 (Unsubstantiated Claims Bias)& 0.250 & 0.667 & 0.111 & 0.467 & 0.333\\
31 (Word Choice Bias)& 0.333 & 0.667 & 0.250 & 0.400 & 0.333
\end{tabular}
\caption{Label-wise agreement statistics for the 11 labels assigned at least once.}
\label{tab:perlabel}
\end{table}

\section{Behaviour Under Controlled Data Generators}

\begin{table}[htbp]
  \centering
  \caption{Representative agreement metrics (averaged over 
           $n\!\in\!\{3,5,10\}$, $N\!\in\!\{20,50,100\}$, 
           $k\!\in\!\{5,10,25\}$, 20 repeats, 100 bootstrap iterations for boot-F1)}
  \label{tab:agreement_metrics_extended}
  \setlength{\tabcolsep}{5.5pt}
  \begin{tabular}{@{}lcccccccc@{}}
    \toprule
    Scenario & $A_{ml}^{(F)}$ & $A_{ml}^{(ac1)}$ & $A_{ml}^{(pb)}$ & MASI & Aug. $\kappa$ & boot-F1 & $A_m$ \\
    \midrule
    Perf. Agr. + High Prev.     & 1.000 & 1.000 & 1.000 & 1.000 &  0.042 &  1.000 & 1.000 \\
    Perf. Agr. + Mod. Prev.     & 1.000 & 1.000 & 1.000 & 1.000 &  0.171 &  1.000 & 1.000 \\
    Perf. Agr. + Low Prev.      & 1.000 & 1.000 & 1.000 & 1.000 &  0.381 &  1.000 & 1.000 \\
    High Agr. + High Prev.      & 0.524 & 0.711 & 0.641 & 0.870 &  0.030 &  0.523 & 0.426 \\
    High Agr. + Mod. Prev.      & 0.634 & 0.647 & 0.641 & 0.817 &  0.101 &  0.619 & 0.495 \\
    High Agr. + Low Prev.       & 0.523 & 0.710 & 0.640 & 0.693 &  0.164 &  0.487 & 0.424 \\
    Indep. + High Prev.         & -0.004 & 0.532 & 0.362 & 0.791 &  0.000 &  0.002 & 0.002 \\
    Indep. + Mod. Prev.         & -0.005 & 0.006 & 0.001 & 0.547 & -0.000 & -0.000 & -0.000 \\
    Indep. + Low Prev.          & -0.007 & 0.529 & 0.359 & 0.354 & -0.000 & -0.001 & -0.001 \\
    Min. Feas. + High Prev.     & -0.287 & 0.230 & 0.052 & 0.687 & -0.017 & -0.274 & -0.113 \\
    Min. Feas. + Mod. Prev.     & -0.287 & -0.155 & -0.215 & 0.468 & -0.033 & -0.263 & -0.076 \\
    Min. Feas. + Low Prev.      & -0.287 & 0.230 & 0.052 & 0.155 & -0.071 & -0.236 & -0.113 \\
    \bottomrule
  \end{tabular}
\end{table}

To study how the three normalizations respond to different data characteristics and how they compare to the measures presented in Section~\ref{related_work}, we generated synthetic annotation matrices. The purpose is descriptive: the simulations show the consequences of the models and do not establish that one coefficient is generally more valid than another.
Two orthogonal factors control the data:

\begin{description}
\item[Prevalence:]
For every label the probability that it is \emph{set} in the
hidden ground-truth vector is fixed to
\emph{high} $(p = 0.80)$, \emph{moderate} $(p = 0.50)$ or
\emph{low} $(p = 0.20)$.

\item[Agreement level:]
After the ground-truth vectors have been drawn, the individual
annotator decisions are created as follows.
\begin{enumerate}
\item \textbf{Perfect agreement} -- every annotator copies the
ground truth, hence the observed agreement
$A_{o}=1.0$.
\item \textbf{High agreement} -- each 0/1 decision is
independently flipped with a small error probability
$\varepsilon = 0.10$.
\item \textbf{Independent draws} -- annotators ignore the
ground truth and draw their own 0/1 decision from the
same $\text{Bernoulli}(p)$ distribution; any agreement
that occurs is therefore implied by the marginal distribution.
\item \textbf{Minimum feasible agreement} -- for every
item--label pair we first compute
$a=\max\bigl(1,
\min(n-1,
\lfloor pn+0.5\rfloor)\bigr)$ and then choose
$a$ of the $n$ annotators at random to assign the value
$1$, while the remaining $n-a$ assign $0$.
This guarantees that the realised prevalence
$\hat{p}=a/n$ matches the requested $p$ (up to the
unavoidable rounding of $a$) and at the same time
forces every pair of annotators to disagree on as many
decisions as possible given that prevalence.
\end{enumerate}
\end{description}

For every combination
$n\in\{3,5,10\}$ annotators,
$N\in\{20,50,100\}$ items and
$k\in\{5,10,25\}$ labels, the simulation was repeated $20$ times;
the table reports the averages of the resulting
$3\times3\times3\times20 = 540$ metric values per scenario.

Table~\ref{tab:agreement_metrics_extended} reports averages over $n$, $N$, and $k$. This compact summary makes the main prevalence interaction visible but should not be read as a study of sampling variance; a full assessment would keep these dimensions separate and report interval estimates.

For perfect agreement, all measures except augmented kappa attain 1.0. The different value of augmented kappa follows from its specific model for multiply labelled data and illustrates why results from different coefficients should not be placed on one interpretive scale without qualification.

Under independent Bernoulli draws, the pooled-marginal coefficient remains close to zero because its baseline matches the generator. AC1 and PABAK become positive at extreme prevalence because they use different baselines. This is not a simulation failure: it shows that zero has a different meaning under the three models. Direct MASI similarity also increases with the number of shared positive labels and is not centred at a chance value.

\begin{figure}[htbp]
    \centering
\includegraphics[width=0.96\linewidth]{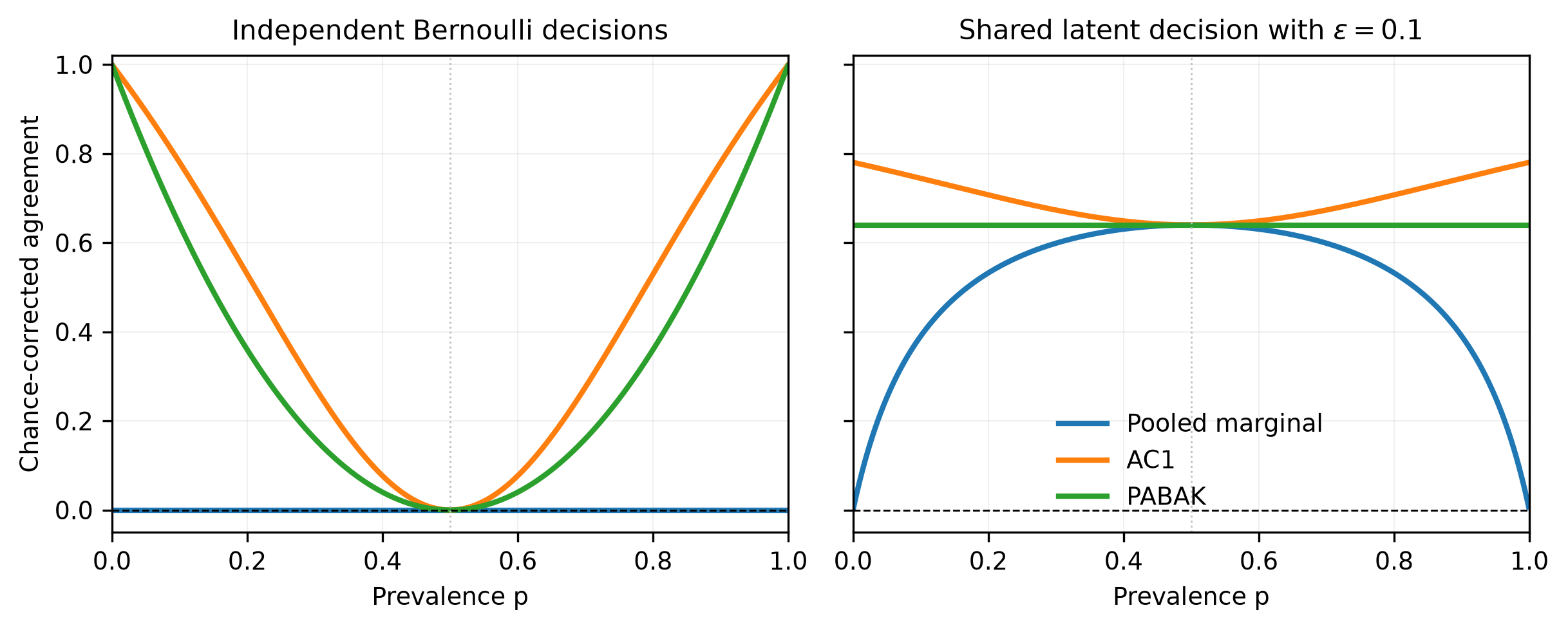}
    \caption{Analytic prevalence sensitivity. Left: annotators draw independently from $\mathrm{Bernoulli}(p)$. Right: annotators copy a latent label and independently flip it with probability $\varepsilon=0.1$. The pooled-marginal coefficient is calibrated to the independent generator; AC1 and PABAK use different baselines.}
    \label{fig:prevalence-curves}
\end{figure}

In the noisy-ground-truth condition, pooled-marginal agreement is smaller at extreme prevalence, AC1 is larger, and PABAK is constant. These patterns can also be obtained analytically. With flip probability $\varepsilon$, two annotators agree with probability $(1-\varepsilon)^2+\varepsilon^2$, while their marginal positive probability is $q=\varepsilon+p(1-2\varepsilon)$. Figure~\ref{fig:prevalence-curves} plots the resulting coefficients continuously over $p$ and also shows the independent-draw generator.

The minimum-feasible-agreement rows also confirm the lower-bound analysis in Equation~\ref{eq:floor}: with more than two annotators and fixed binary marginals, strongly negative values may be unattainable. This is why universal verbal cut-offs such as ``good'' or ``poor'' agreement are particularly inappropriate here.

The code used to generate Figure~\ref{fig:prevalence-curves} is included with the preprint source. The Monte Carlo implementation is available in the accompanying software repository.

\section{Reporting Recommendations}

The analysis suggests that a multi-label reliability report should make the following choices explicit:
\begin{enumerate}
\item \textbf{Label universe.} State whether the inventory was fixed before annotation and whether every annotator was expected to consider every label. Never-selected labels should not be removed only because they lower or raise a preferred summary.
\item \textbf{Meaning of joint omissions.} Use Hamming-based agreement when joint absence is informative. If overlap among positive selections is the scientific focus, additionally report a set distance such as Jaccard or MASI, preferably within a chance-corrected framework when such correction is needed.
\item \textbf{Baseline model.} Pre-specify the primary chance model. A pooled-marginal coefficient, AC1, and PABAK do not provide three interchangeable estimates of the same quantity.
\item \textbf{Diagnostics.} Report $N$, the number of annotators per item, $k$, $A_o$, label-wise prevalences and agreements, and the conditional lower bound. Item-level values can identify difficult examples, but reliability conclusions require an adequate number of items.
\item \textbf{Uncertainty.} For substantive studies, accompany corpus-level and important label-level estimates with confidence intervals obtained through a design-appropriate bootstrap or an analytic method. Resampling must respect the item and annotator structure of the study.
\end{enumerate}
If labels are hierarchical, mutually exclusive in groups, or otherwise dependent, the additive Hamming construction is often not sufficient. In that case, Krippendorff's framework with a domain-specific distance, or an explicit probabilistic annotator model, is preferable to treating the dependencies as if they did not exist.

\section{Implementation}

A Python implementation is available in the \href{https://github.com/Timperator2/MultiLabelAgreement}{project repository} and from the Python Package Index as \href{https://pypi.org/project/ml-agreement/}{\texttt{ml-agreement}}. The package implements corpus-, item-, and label-level summaries, the three baseline choices discussed here, label prevalence, and the conditional minimum-agreement calculation. The pooled-marginal option is retained as the package's legacy \texttt{k}/kappa option; in the paper we use $F$ to avoid confusing pooled marginals with Cohen's annotator-specific model.

\section{Limitations}

The present analysis is restricted to complete, balanced annotation designs with binary, non-hierarchical labels. The simple formulas do not by themselves handle missing annotations. Treating a missing decision as 0 would incorrectly turn missingness into an observed omission; instead, pair counts and marginal denominators must be computed from the available values, as done in the general formulation of Krippendorff's $\alpha$.

Normalized Hamming distance is additive and therefore insensitive to dependencies among labels. It also weights all label decisions equally before chance correction. Hierarchical relations, mutually exclusive subsets, and differences in the cost of false positive and false negative decisions require a different distance or a model tailored to the annotation design.

The worked example contains only four items and is illustrative. The simulation varies several design factors but averages over $n$, $N$, and $k$ and does not provide inferential comparisons or confidence intervals. A more complete empirical study should analyse these factors separately, include annotator-specific marginals and correlated labels, and compare against Hamming $\alpha$ using the same data. Finally, tasks with genuine ambiguity may be better represented by confidence judgements or distributions over interpretations rather than binary decisions alone.

\section{Conclusion}

Multi-label agreement does not require an entirely new coefficient family. Average pairwise Hamming agreement, combined with a pooled-marginal chance term, is a chance-disagreement-weighted average of label-wise Fleiss/Scott-type coefficients and is closely related to Krippendorff's $\alpha$ with Hamming distance. Stating this equivalence is useful because it reveals both the strengths and the limitations of the construction: it scales linearly in the number of labels and counts joint omissions, but it decomposes over labels and does not model their dependencies.

The choice of baseline remains consequential. The pooled-marginal, AC1, and fixed $0.5$ baselines can give very different values on the same annotations, especially under extreme prevalence. The simulations show that these differences follow from the models rather than from numerical instability alone. Reporting several coefficients can be a useful sensitivity analysis, but it does not remove the need to select and justify a primary estimand.

The conditional minimum-agreement formula adds a complementary diagnostic. It shows how low observed and normalized agreement can feasibly fall once the number of annotators and empirical label prevalences are fixed. Together with raw agreement, label-wise results, and an explicit declaration of the label universe, this provides a more informative account than a single coefficient. The accompanying implementation packages these calculations so that the assumptions can be inspected rather than hidden behind one recommended score.


\bibliographystyle{compling}
\bibliography{COLI_template}

\end{document}